\documentclass[sigconf]{acmart}
\usepackage{graphicx}
\usepackage{subcaption} 
\usepackage{hyperref}
\usepackage{cleveref}
\usepackage{multicol}
\usepackage{multirow}
\usepackage{afterpage}
\usepackage{amsmath}
\usepackage{tabularx}
\usepackage{float}
\usepackage{dblfloatfix}
\crefname{algocf}{algorithm}{algorithms}
\Crefname{algocf}{Algorithm}{Algorithms}
\usepackage[linesnumbered,ruled,vlined]{algorithm2e}

\AtBeginDocument{%
  }

\copyrightyear{2024}
\acmYear{2024}
\setcopyright{acmlicensed}
\acmConference[KDD '24] {Proceedings of the 30th ACM SIGKDD Conference on Knowledge Discovery and Data Mining }{August 25--29, 2024}{Barcelona, Spain.}
\acmBooktitle{Proceedings of the 30th ACM SIGKDD Conference on Knowledge Discovery and Data Mining (KDD '24), August 25--29, 2024, Barcelona, Spain}
\acmISBN{979-8-4007-0490-1/24/08}
\acmDOI{10.1145/3637528.3671980}

\acmSubmissionID{1837}
\begin{document}

\title{Dynamic Neural Dowker Network: Approximating Persistent Homology in Dynamic Directed Graphs}

\author{Hao Li}
\orcid{0009-0009-9322-0603}
\affiliation{%
  \institution{Wuhan University}
  \department{Electronic Information School}
  \city{Wuhan}
  \state{Hubei}
  \country{China}
  \postcode{430072}
}
\email{whulh@whu.edu.cn}

\author{Hao Jiang}
\orcid{0009-0006-5853-5544}
\authornote{Corresponds Author}
\affiliation{%
\institution{Wuhan University}
\department{Electronic Information School}
  \city{Wuhan}
  \state{Hubei}
  \country{China}
  \postcode{430072}
}
\email{jh@whu.edu.cn}

\author{Fan Jiajun}
\orcid{0000-0002-5003-356X}
\affiliation{%
\institution{Wuhan University}
\department{Electronic Information School}
  \city{Wuhan}
  \state{Hubei}
  \country{China}
  \postcode{430072}
}
\email{2017301200224@whu.edu.cn}

\author{Dongsheng Ye}
\orcid{0000-0001-7042-2581}
\affiliation{%
\institution{Wuhan University}
\department{Electronic Information School}
  \city{Wuhan}
  \state{Hubei}
  \country{China}
  \postcode{430072}
}
\email{YesDong@whu.edu.cn}

\author{Liang Du}
\orcid{0009-0001-4847-4075}
\affiliation{%
  \institution{Wuhan University}
  \department{Electronic Information School}
  \city{Wuhan}
  \state{Hubei}
  \country{China}
  \postcode{430072}
}
\email{duliang@whu.edu.cn}

\renewcommand{\shortauthors}{Hao Li et al.}

\begin{abstract}
  Persistent homology, a fundamental technique within Topological Data Analysis (TDA), captures structural and shape characteristics of graphs, yet encounters computational difficulties when applied to dynamic directed graphs. This paper introduces the Dynamic Neural Dowker Network (DNDN), a novel framework specifically designed to approximate the results of dynamic Dowker filtration, aiming to capture the high-order topological features of dynamic directed graphs. Our approach creatively uses line graph transformations to produce both source and sink line graphs, highlighting the shared neighbor structures that Dowker complexes focus on. The DNDN incorporates a Source-Sink Line Graph Neural Network (SSLGNN) layer to effectively capture the neighborhood relationships among dynamic edges. Additionally, we introduce an innovative duality edge fusion mechanism, ensuring that the results for both the sink and source line graphs adhere to the duality principle intrinsic to Dowker complexes. Our approach is validated through comprehensive experiments on real-world datasets, demonstrating DNDN's capability not only to effectively approximate dynamic Dowker filtration results but also to perform exceptionally in dynamic graph classification tasks. 
\end{abstract}




\maketitle

\section{Introduction}
In recent years, an increasing number of researchers are focusing on integrating high-order topological features with graph learning for downstream tasks such as node classification, link prediction, and graph classification \cite{interactive, tgnn, topoattn}. As a method under the framework of TDA, persistent homology captures multi-scale features of graphs to describe their structural and shape characteristics \cite{filtrationlearning, tgnn, treph}. Nevertheless, when the subject of study is dynamic directed graphs commonly found in the real world, existing methods tailored for undirected static graphs are inadequate in capturing the topological information of graphs.

\begin{figure*}[ht]
  \centering
    \begin{subfigure}{0.4\textwidth}
    \includegraphics[width=\textwidth]{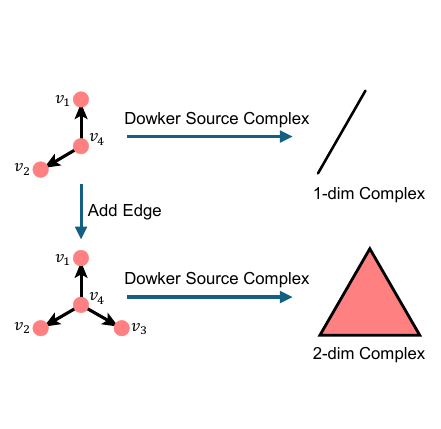}
    \caption{}
    \label{fig1a}
  \end{subfigure}\hfill
  \begin{subfigure}{0.6\textwidth}
    \includegraphics[width=\textwidth]{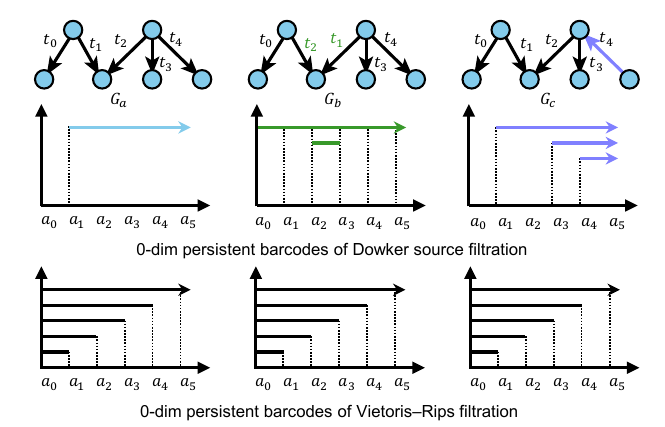}
    \caption{}
    \label{fig1b}
  \end{subfigure}
  \caption{\textbf{(a)} A simple example of a Dowker source complex. The existence of a shared neighbor (in this case, \(v_4\)) between \(v_1\) and \(v_2\) creates a higher-order relationship, which the Dowker source complex captures. \textbf{(b)} Illustration of Dowker filtration sensitive to edge weights and directions in graphs: \(G_a\) represents a common type of subgraph in social media diffusion graphs. Swapping the timestamps of two edges (\(G_b\)) or changing the direction of an edge (\(G_c\)) leads to different diffusion. Dowker source filtration can effectively distinguish these three types of graphs, whereas Vietoris-Rips (VP) filtration generates the same persistent barcode for all three cases. }
  \label{fig1}
\end{figure*}

Dynamic Dowker filtration is an effective persistent homology method for capturing the structural and shape characteristics of dynamic directed graphs \cite{SDPH, lowrank, dihomo}. As demonstrated in \cref{fig1a}, Dowker complexes focus on capturing the shared neighbor structure of graphs. This approach is particularly sensitive to the direction and weight of edges, a feature illustrated with a simple example in \cref{fig1b}. Such sensitivity makes Dowker complexes adept at analyzing complex, continuously evolving dynamic directed graphs in the real world. Their ability to discern nuanced relationships based on edge directionality and weights allows for a more detailed understanding of these graphs' topological characteristics.

Similar to other persistent homology methods, Dowker complexes face computational challenges, struggling to efficiently handle dynamic graphs. The work in \cite{canper} explores the capability of neural networks to learn persistent homology features in digital images and filtered cubical complexes. Inspired by graph neural executors \cite{neuralexecution, transfer}, we aim to develop a learning-based method to approximate the complex results of Dowker computations. Graph neural executors are a type of neural network designed to approximate the execution of algorithms on graphs. The literature \cite{neuralapproximation} has utilized graph neural networks to approximate extended persistence diagrams (EPDs), demonstrating that persistent homology results can be effectively predicted through embeddings of edges. 

However, applying graph neural executors to Dowker dynamic filtration presents two key challenges. \textbf{(1) How to approximate with a neural network the structural features captured by Dowker complexes.} Traditional graph neural networks, relying on information transfer between a node and its neighbors, are not adept at directly representing the computational results of Dowker complexes. \textbf{(2) How to preserve the inherent duality characteristic of Dowker complexes.} Dowker complexes, designed for directed graphs, can be divided into source Dowker complexes and sink Dowker complexes \cite{methods}. According to the principle of duality, under the same filtration, the Persistence Diagrams (PDs) corresponding to both types of complexes should be identical \cite{homologygroup}. It is imperative for the neural executor to ensure consistency between these two forms of complexes. This involves designing a neural network architecture that can simultaneously capture and reconcile the distinct yet complementary information presented in the source and sink Dowker complexes, reflecting their dual nature in the PDs.

\begin{figure}[ht]
  \centering
  \includegraphics[width=\columnwidth]{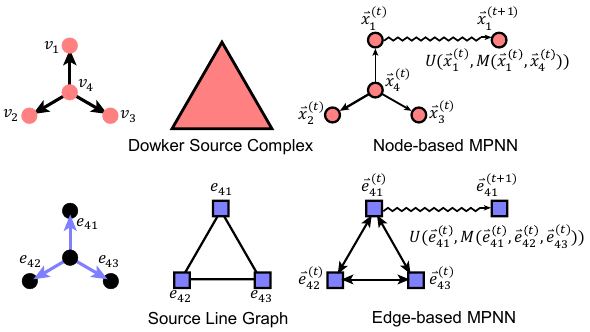}
  \caption{An example demonstrating the relationship between the source Dowker complex and the source line graph for a directed graph.  In the source line graph, the presence of an edge between nodes representing \(e_{41}\) and \(e_{42}\) reflects the shared neighbor (\(v_4\)) between \(v_1\) and \(v_2\) in the original graph. This edge-based perspective is particularly allows the neural network to focus on the interactions and relationships between edges.}
  \label{fig2}
\end{figure}

To address the aforementioned challenges, we focus on line graphs \cite{linegraph, lgnn, lwgnn}. A line graph \(L(G)\) transforms the edges of a graph \(G\) into nodes of a new graph and connects edges that have a common node in \(G\). As depicted in \cref{fig2}, the line graph establishes a critical linkage between Dowker complexes and Graph Neural Networks (GNNs), facilitating the direct computation of edge embeddings essential for predicting persistent homology. Further, this paper expands on the definition of directed line graphs. Based on the direction of the edges, a directed graph is transformed into source and sink line graphs, corresponding to Dowker complexes. This approach aligns well with the duality requirement of Dowker complexes.

Specifically, we develop the Dynamic Neural Dowker Network (DNDN), designed to approximate the computational results of dynamic Dowker filtration and further validated its effectiveness in dynamic graph classification tasks. Initially, we employ a line graph transformation method to convert the original directed graph into both source and sink line graphs, capturing the intricate edge transformation processes in dynamic  directed graphs. Subsequently, we utilize a Source-Sink Line Graph Neural Network (SSLGNN) to capture the neighbor structures of dynamic edges, aiming to align with the computational outcomes of dynamic Dowker filtration and adapt to the evolving nature of dynamic graphs. Within the SSLGNN framework, we introduce a duality edge fusion mechanism, ensuring that the neural execution results for both the sink and source line graphs meet the duality requirements of Dowker complexes. In essence, our main contributions can be summarized as follows:

\begin{itemize}
  \item We proposed an innovative line graph transformation method, which creates sink and source line graphs to directly represent the shared neighbor structures of interest in Dowker complexes. 
  \item We developed the Dynamic Neural Dowker Network (DNDN), leveraging Source-Sink Line Graph Neural Network to analyze complex high-order structures in dynamic directed graphs. Additionally, we designed a duality edge fusion mechanism to align with the unique properties of Dowker complexes.
  \item We conducted experiments on real-world datasets and found that DNDN performs well in approximating the computational results of dynamic Dowker filtration and in graph classification tasks. These results demonstrate the effectiveness of our approach in practical applications.
\end{itemize}

\section{Related Works}
In this section, we discuss the relevant research on dynamic graph neural networks and persistent homology with graph learning.

\subsection{Dynamic graph neural networks}
Recent studies on dynamic graphs mainly rely on temporal granularity and are divided into two types: discrete-time dynamic graphs using time snapshots and continuous-time dynamic graphs with edge timestamps \cite{dynamicsurvey}. Existing research on dynamic graph neural networks focuses on extending traditional static GNNs to dynamic graphs. An intuitive approach involves using time-aware encoders to capture the dynamic evolution of nodes between discrete snapshots. For example, EvolveGCN \cite{evolvegcn} dynamically updates the GNN model weights using LSTM and GRU. DySAT \cite{dysat} and DHGAT \cite{dhgat} employ attention mechanisms to capture node embeddings through structural and temporal evolution. Recent dynamic GNNs based on meta-learning aim to model temporal factors, with meta-learning adeptly adapting to new temporal data. The Roland framework \cite{roland} generates new node embeddings through meta real-time updates. WinGNN \cite{wingnn} utilizes a meta-learning strategy to model associations in sliding windows of adjacent and consecutive snapshots, without the need for specific time-aware encoders. However, these dynamic network studies, based on neighborhood aggregation GNN architectures, struggle to capture the global topological characteristics of dynamic graphs from a holistic perspective. Our approach employs graph neural networks to approximate the results of dynamic Dowker filtration, obtaining global topological features of dynamic graphs and providing a new perspective on understanding the evolution of dynamic graphs.

\subsection{Persistent Homology with Graph Learning}
As a significant component in topological machine learning, the integration of persistent homology with graph learning has garnered widespread attention. 
Given the superiority of persistent homology in capturing the global structural information of graphs, a considerable amount of research has focused on predicting the labels of entire graphs, that is, graph classification tasks \cite{barcodes, filtrationlearning, treph, perslay, topoattn, toposignatures, beyond}. Some methods \cite{enhanced, trignn} utilize the topological features of node neighborhood subgraphs as representations for node classification tasks. Similarly, \cite{interactive} constructs neighborhood graphs for each pair of target nodes and calculates their topological features, applying the outcomes to link prediction tasks. TOGL \cite{tgnn} developed a universal layer that integrates topological information into the hidden representations of nodes, capable of computing topological features at all scales, and demonstrated the efficacy of this approach in both graph classification and node classification tasks.

Due to the high computational complexity of traditional persistent homology calculations, some methods have focused on using neural networks to approximate the results of persistent homology, especially for large-scale graphs. Inspired by neural executors \cite{neuralexecution}, the work in \cite{neuralapproximation} reinterprets the computation of extended persistent homology as a prediction problem of paired edges, which can be resolved using a union-find algorithm. Subsequently, a graph neural network is designed to learn this union-find algorithm, and the effectiveness of the resulting Edge-based Persistence Diagrams (EPDs) in downstream tasks has been validated.

These studies successfully integrate the topological features of graphs with graph learning methods. However, most of these approaches focus predominantly on static, undirected graphs, thereby overlooking the rich and complex information present in real-world directed dynamic graphs. Addressing this gap, our paper introduces the Dynamic Neural Dowker Network, which focuses on machine learning methods for computing persistent homology, an area less explored in existing work. We combine edge-based Dowker complexes with neural networks to execute persistence diagram (PD) computations and apply this approach to dynamic graph classification tasks. This innovation extends the application of topological data analysis to more complex and dynamic network structures.

\section{PRELIMINARY AND BACKGROUND}
In this section, we first provide a brief overview of the concepts that are relevant to persistent homology. We then describe the Dowker filtration on graphs, and finally, we discuss the concepts related to line graphs.

\subsection{Persistent Homology}

Persistent homology is a key technique in topological data analysis, which studies the shape and structure of data. The central idea is to analyze datasets by constructing a series of topological spaces and observing the persistence of features across changes in a parameter. When focusing on persistent homology in the context of graphs, the approach typically involves examining how graph-based structures evolve.

Given a graph \(\mathcal{G}=(\mathcal{V}, \mathcal{E})\), where \(\mathcal{V}\) is the set of nodes and \(\mathcal{E}\) is the set of edges, with \(e_{m n} \in \mathcal{E}\) representing an edge between nodes \(v_{m}, v_{n} \in \mathcal{V}\). The key step in computing persistent homology involves creating a nested sequence of subgraphs \(\mathcal{G}_1 \subseteq \mathcal{G}_2 \subseteq \ldots \subseteq \mathcal{G}_k=\mathcal{G}\) based on a filter function \(f\), let \(\mathcal{C}_i\) be the simplicial complex induced by \(\mathcal{G}_i\). This nested sequence of simplicial complexes \(\mathcal{C}_1 \subseteq \mathcal{C}_2 \subseteq \ldots \subseteq \mathcal{C}_k\) is referred to as a filtration of \(\mathcal{G}\).

Using the additional information provided by the filtration, we can obtain the persistent homology groups \(H\) and their ranks \(\beta\). The \(\beta_{k}^{i,j}\) captures the topological features of the graph at various dimensions \( k \), such as connected components for \(k=0\) and loops for \(k=1\), \( i \) and \( j \) represent the birth and death of these topological features. The computed persistent homology can be encoded as a multi-point set in \( R^2 \), known as the persistence diagram (PD), where the \(x\) and \(y\) coordinates represent the birth and death of topological features, respectively. This approach to analyzing graphs captures their underlying topological characteristics, providing valuable insights into their structure and dynamics beyond what is observable through traditional graph metrics.

\subsection{Dowker Filtration}
The Dowker Complex is a type of simplicial complex specifically designed for directed graphs. Given a weighted directed graph \(\mathcal{G}_d=(\mathcal{V}_d, \mathcal{E}_d, \mathcal{W}_d)\), where \(\mathcal{V}_d\) is the set of nodes, \(\mathcal{E}_d\) is the set of directed edges, and \(e_{m n} \in \mathcal{E}_d\) represents a directed edge from source node \(v_{m}\) to target node \(v_{n}\), with \(\mathcal{W}_d(e_{m n})\) being the weight of the edge \(e_{m n}\). The Dowker \(\delta\)-sink complex is defined as a simplicial complex as follows:

\begin{align}
  \label{equ1}
  \mathfrak{D}_{\delta, \mathcal{G}}^{s i} := \{ & \sigma=\left[v_0, \ldots, v_n\right]: \text{ there exists } v_a \in \mathcal{V} \nonumber \\
  & \text{ such that } \mathcal{W}\left(e_{b a}\right) \leq \delta \text{ for each } v_b \in \{v_0, \ldots, v_n\} \}.
\end{align}

The node \(v_a\) is defined as a \(\delta\)-sink of a simplex \(\sigma\) if, for all nodes \(v_b\) in \(\sigma\), there is a directed edge \(e_{b a}\) from $v_b$ to \(v_a\) with a weight less than or equal to \(\delta\).

Similarly, the Dowker \(\delta\)-source complex is symmetrically defined with the roles of source and target reversed:
\begin{align}
  \label{equ2}
  \mathfrak{D}_{\delta, \mathcal{G}}^{s o} := \{ & \sigma=\left[v_0, \ldots, v_n\right]: \text{ there exists } v_a \in \mathcal{V} \nonumber \\
  & \text{ such that } \mathcal{W}\left(e_{a b}\right) \leq \delta \text{ for each } v_b \in \{v_0, \ldots, v_n\} \}.
\end{align}

With increasing \(\delta\), we naturally obtain a sequence of Dowker complexes, known as the Dowker \(\delta\)-sink filtration or Dowker \(\delta\)-source filtration. This filtration process allows for the examination of the evolving topological structure of the graph as the parameter \(\delta\) changes, providing insights into the hierarchical and directional properties of the graph. 
\begin{align}
  \left\{\mathfrak{D}_\delta^{s i} \hookrightarrow \mathfrak{D}_{\delta^{\prime}}^{s i}\right\}_{0 \leq \delta \leq \delta^{\prime}},  \\
  \left\{\mathfrak{D}_\delta^{s o} \hookrightarrow \mathfrak{D}_{\delta^{\prime}}^{s o}\right\}_{0 \leq \delta \leq \delta^{\prime}}.
\end{align}

According to \cite{SDPH}, by converting the temporal information carried by edges into weights, we can achieve a stable \textbf{dynamic Dowker filtration}. Given the sensitivity of Dowker filtration to edge directionality and weights, dynamic Dowker filtration is capable of deeply analyzing the high-order interaction relationships in dynamic graphs as they evolve over time. This approach leverages the inherent temporal dynamics of edges, allowing for a more nuanced understanding of how graphs change and how these changes impact the topological features of interest.

\textbf{Dowker Duality.} For a given directed graph \(\mathcal{G}_d\), any threshold value \(\delta \in \mathbb{R}\), and dimension \(k\geq 0\), the persistent modules induced by the Dowker sink filtration and the Dowker source filtration are isomorphic. This statement reflects a fundamental property of Dowker complexes in topological data analysis. The duality of Dowker complexes necessitates that, in designing neural networks, we must consider both the distinctions and the eventual consistency between source Dowker complexes and sink Dowker complexes.

\subsection{Line Graph}

Given an undirected graph $\mathcal{G}=(\mathcal{V}, \mathcal{E})$ with $E \neq \emptyset $, the corresponding line graph $L(\mathcal{G})=(\mathcal{V}_{L(\mathcal{G})},\mathcal{E}_{L(\mathcal{G})}) $ is defined as follows:

\begin{align} 
  \mathcal{V}_{L(\mathcal{G})} =& \{v_e | e \in \mathcal{E}\}, \\
  \mathcal{E}_{L(\mathcal{G})} =& \{\{v_e, v_f\} | e \neq f  \text{ and } e \cap f \neq \emptyset\},
\end{align}

The threshold \(\delta\) in Dowker filtration is defined on the edges, which contrasts with the node-centric approach common in traditional graph neural networks. To address this, our paper extends the concept of line graphs, introducing sink line graphs and source line graphs, corresponding to Dowker complexes. Specifically, given a weighted directed graph \(\mathcal{G}_d=(\mathcal{V},\mathcal{E} )\), its corresponding source line graph \(L(\mathcal{G}_d)^{so}=(\mathcal{V}_{L(\mathcal{G}_d)}^{so}, \mathcal{E}_{L(\mathcal{G}_d)}^{so}) \) and sink line graph \( L(\mathcal{G}_d)^{si}=(\mathcal{V}_{L(\mathcal{G}_d)}^{si}, \mathcal{E}_{L(\mathcal{G}_d)}^{si}) \)  are defined as follows:

\begin{align} 
  \mathcal{V}_{L(\mathcal{G}_d)}^{so} &= \mathcal{V}_{L(\mathcal{G}_d)}^{si} = \{v_{e} | e \in \mathcal{E}\}, \\
  \mathcal{E}_{L(\mathcal{G}_d)}^{so} &= \{(v_a, v_b) | a \neq b , \text{ and } source(a) = source(b) \},\\
  \mathcal{E}_{L(\mathcal{G}_d)}^{si} &= \{(v_a, v_b) | a \neq b , \text{ and } target(a) = target(b) \},
\end{align}

where \(source(a)\) and \(target(a)\) respectively refer to the source and target nodes of an edge \(a\), \(L(\mathcal{G}_d)^{so} \) and \(L(\mathcal{G}_d)^{si} \), are defined as undirected graphs.This adaptation allows the integration of edge-centric filtration methods into graph neural network frameworks, suitable for analyzing dynamic directed graphs.

\section{method}
\begin{figure*}[ht]
  \centering
  \includegraphics[width=0.9\textwidth]{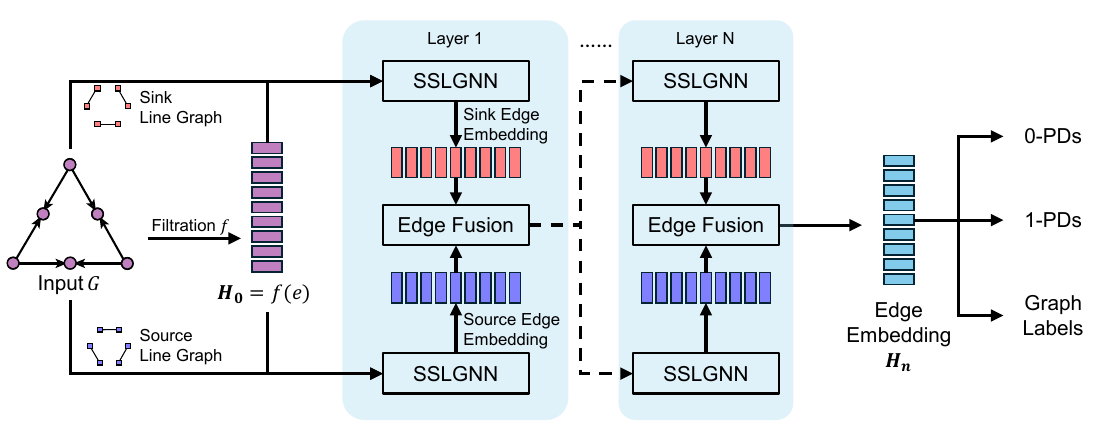}
  \caption{The framework of DNDN}
  \label{fig3}
\end{figure*}

Given a dynamic directed graph $ \mathcal{G}^t=(\mathcal{V}^t, \mathcal{E}^t) $, our goal is to learn a function $ F: \mathcal{E}^t \rightarrow \mathbb{R}^2 $ to approximate the 0-dimensional and 1-dimensional PDs under dynamic Dowker filtration. To predict the high-order topological features of dynamic graphs, we propose the overall framework of the Dynamic Neural Dowker Network (DNDN) as shown in \cref{fig3}. It specifically includes the following modules: (1) Line Graph Embedding Module. The DNDN obtains embeddings for each edge in the dynamic graph through a source-sink line graph neural network combined with an edge fusion layer. (2) Joint Prediction Module. The DNDN divides the prediction task into two joint tasks: PD prediction and graph label prediction, and designs the loss function incorporating the Wasserstein distance.

\subsection{Line Graph Embedding Module}
To obtain edge embeddings for predicting persistent homology under dynamic Dowker filtration, this paper utilizes line graph transformations to derive the source and sink line graphs corresponding to the dynamic directed graphs. Two distinct line graph neural networks are employed to calculate embeddings for the source and sink line graphs, respectively. Finally, an edge fusion mechanism is implemented to ensure the duality of Dowker complexes.

\textbf{Line graph transformation.} Initially, for the given input dynamic directed graph \(\mathcal{G}^t=(\mathcal{V}^t, \mathcal{E}^t)\),  we transform it into source and sink line graphs according to \cref{equ1} and \cref{equ2}. The temporal attributes of \(\mathcal{E}^t\) are transferred to the nodes in the line graphs \(\mathcal{V}_{L(\mathcal{G}^t)}^{si}\) and \(\mathcal{V}_{L(\mathcal{G}^t)}^{so}\), facilitating subsequent processing by the source-sink line graph neural network.

\textbf{Source-sink line graph neural network.} Next, we introduce the backbone neural network layer used for generating edge embeddings in line graphs. After the transformation, the original directed graph is divided into source line graphs and sink line graphs. We have designed a Source-Sink Line Graph Neural Network (SSLGNN) to generate their edge embeddings. The SSLGNN consists of two distinct line graph neural networks and an edge fusion module. Specifically, the computation process of the $m$-th layer of SSLGNN is as follows:
\begin{align}
  h_{u_{so}}^m =& A G G^m\left(\left\{M S G^m\left(h_{v_{ef}}^{m-1}\right), v_{so} \in \mathcal{N}(u_{so})\right\}, h_{u_{ef}}^{m-1}\right) \\
  h_{u_{si}}^m =& A G G^m\left(\left\{M S G^m\left(h_{v_{ef}}^{m-1}\right), v_{si} \in \mathcal{N}(u_{si})\right\}, h_{u_{ef}}^{m-1}\right) \\
  h_{u_{ef}}^m =& EdgeFusion \left( h_{u_{so}}^m, h_{u_{si}}^m\right)
  \end{align}

where \(h^m_{u_{so}}\) and \(h^m_{u_{si}}\) represent the features of edge \(u\) at the \(m\)-th layer in the source line graph and sink line graph, respectively, while \(h^m_{u_{ef}}\) denotes the features after edge fusion. \(\mathcal{N}(u_{so})\) and \(\mathcal{N}(u_{si})\) respectively denote the neighborhoods of edge \(u\) in the source line graph and sink line graph. \(AGG\) represents the aggregation function, and \(MSG\) denotes the message-passing function. The configurations of \(AGG\) and \(MSG\) are consistent with those described in \cite{neuralapproximation}.

The initial features of an edge \(h^0_{u_{so}} = h^0_{u_{si}}\) are computed using a dynamic Dowker filtration. We have developed a simple yet efficient dynamic Dowker filtration \(\mathcal{W}(\cdot )\) to meet both the requirements of dynamic Dowker duality and dynamic Dowker structural stability \cite{SDPH}. The initial feature calculation is defined as follows:

\begin{equation}
  h^0_{u_{so}} = h^0_{u_{si}} = \mathcal{W}(u) = \frac{t_u - t_{min}}{t_{max} - t_{min}},
\end{equation}

where \(t_{u}\) represents the time at which the edge \(u \) appears, and \(t_{min}\) and \(t_{max}\) are respectively the minimum and maximum values among all edge appearance times \(\mathcal{T}\) in the dynamic graph.

\subsection{Joint Prediction Module}
The topological features at different dimensions of a graph represent various structural aspects, such as connected components for \(k=0\) and loops for \(k=1\). However, the quantity of Dowker PDs in a graph does not directly correlate with the number of nodes or edges. To better characterize topological features across dimensions, we have designed a prediction module for simultaneously predicting the 0-dimensional and 1-dimensional Persistence Diagrams (0-PDs and 1-PDs) of dynamic graphs. Simultaneously, inspired by the concept of joint learning, we have designed a graph label prediction module to accommodate the needs of downstream tasks. 

\textbf{0-PD Prediction.}
We extend the concept from \cite{treph} and classify the elements of a graph's 0-PD into three elements:
\begin{enumerate}
  \item Paired points $(a, b)$, representing a connected component born at $a$ and dying at $b$ (merging into a higher-dimensional topological structure).
  \item Unpaired points $(a, +\infty)$, representing a connected component that is born but does not die.
  \item Disappearing points $(a, a)$, indicating that the corresponding connected component quickly dies after formation.
\end{enumerate}

\begin{figure}[ht]
  \centering
  \includegraphics[width=0.9\columnwidth]{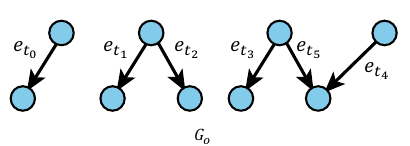}
  \caption{An example demonstrating the 0-PD of Dowker complexes.}
  \label{fig4}
\end{figure}

As shown in \cref{fig4}, in graph \(G_o\), with \(t_0 < t_1 < t_2 < t_3 < t_4$, $(t_0, +\infty)\), \((t_1, +\infty)$, $(t_3, +\infty)\) correspond to the unpaired points for \(e_{t_0}, e_{t_1}, e_{t_3}\) respectively, \((t_4, t_5)\) corresponds to the paired point for \(e_{t_4}\), and the edges \(e_{t_2}, e_{t_5}\) forming higher-dimensional complexes correspond to the disappearing points \((t_2, t_2), (t_5, t_5)\). This approach directly correlates edges with 0-PDs without altering the true computational results. Furthermore, we can utilize an MLP layer to directly predict the PD corresponding to an edge.

\begin{equation}
  PD^0 = MLP(\mathbf{H^n}),
\end{equation}
where $ PD^0 $ represents the predicted results for 0-dimensional Persistence Diagrams (0-PD), and $ H^n $ denotes the edge embeddings output by the Source-Sink Line Graph Neural Network (SSLGNN). 

\textbf{1-PD Prediction.} The interpretation of the birth and death of 1-dimensional Dowker PDs from a geometric perspective is challenging, as each PD point is associated with multiple edges. Therefore, we propose a neighborhood-based aggregation method that utilizes dynamic Dowker filtration values \(\mathcal{W}(\cdot)\) to weight the aggregation of each edge's neighborhood, resulting in a subgraph structural representation. Subsequently, a Multilayer Perceptron (MLP) layer is employed to calculate the 1-PD.

\begin{align}
  h^n_{\text{agg}} &= \sum_{v \in \mathcal{N}(u_{si}) \bigcup \mathcal{N}(u_{so})} \mathcal{W}(v) \cdot h^n_v, \\
  PD^1 &= MLP(\mathbf{H}^n_{\text{agg}}),
\end{align}
in this formulation, \(h^n_v \) belongs to \(\mathbf{H}^n \) and the neighborhood of edge \(u\), denoted as \(\mathcal{N}(u_{si}) \cup \mathcal{N}(u_{so})\), includes its neighborhoods in both the source and sink line graphs.

\textbf{Graph Label Prediction.} We propose a graph classification module based on the idea of joint learning, which predicts the labels of graphs after pooling the edge embeddings.

\begin{equation}
    \hat{y} = f(\text{Pooling}(\mathbf{H}_n))
\end{equation}
where \(\hat{y}\) represents the predicted label of the graph, and \(\text{Pooling}\) is the pooling operation, for which we opt for max pooling in our method.

For the PD prediction task, we employ the 2-Wasserstein distance between the predicted results and the ground truth as the loss function. Simultaneously, for the graph classification task, we choose cross-entropy as the loss function to train the model. This dual approach ensures that our model is not only capable of accurately approximating PDs but also effectively classifies graphs, leveraging the topological features captured by the PDs for enhanced performance in classification tasks.

\section{experiments}
In this section, we provide a thorough evaluation of the proposed Dynamic Neural Dowker Network from three perspectives. \Cref{5.1} details the datasets and baselines employed in our experiments. In \Cref{5.2}, we assess the model's proficiency in approximating true persistent homology outcomes. \Cref{5.3} evaluates the model's transferability and efficiency. Building upon the insights gained from the initial experiments, \Cref{5.4} is dedicated to validating the model's performance in graph classification tasks. The results demonstrate that our algorithm successfully approximates the computational results of Dowker complexes and can be effectively applied to downstream graph classification tasks, with notable efficiency and transferability to larger graphs.\footnote{The source code of DNDN is available at \url{https://github.com/Lihaogx/DNDN}}

\subsection{Datasets and baselines}\label{5.1}
The datasets used in the experiments are categorized into two types: static and dynamic, with both large and small graph datasets constructed for each category. The static datasets include REDDIT-BINARY, REDDIT-MULTI-5K and REDDIT-MULTI-12K, where Reddit serves as an online forum, with nodes representing users and edges representing discussion threads. The dynamic datasets encompass four categories: citation graphs, Bitcoin graphs, Q$\&$A graphs and social graphs. The Bitcoin graphs consist of who-trusts-whom graphs from two Bitcoin platforms: Bitcoin OTC and Bitcoin Alpha. The citation graphs include two distinct domains of paper citation graphs: HEP-PH (high energy physics phenomenology) and HEP-TH (high energy physics theory). The Q$\&$A graphs compile records from various websites, including StackOverflow, MathOverflow, SuperUser, and AskUbuntu. The social graph datasets consist of Hashtag diffusion graphs from the Weibo platform, categorized into two types: entertainment and current affairs topics. \Cref{tab1} displays the details of these datasets.

\begin{table}[h!]
  \centering
  \caption{Statistics of the datasets}
  \label{tab1}
  \small
  \setlength{\tabcolsep}{3pt}
  \begin{tabular}{lccccccc} 
    \toprule
    \multirow{3}{*}{Datasets} & & \multicolumn{3}{c}{Smail Graphs} & \multicolumn{3}{c}{Large Graphs} \\
     \cmidrule(lr){3-5} \cmidrule{6-8} 
    & \multirow{2}{*}{Classes} & \multirow{2}{*}{Graphs} & Avg. & Avg. & \multirow{2}{*}{Graphs} & Avg. & Avg. \\ 
    & & &  Nodes & Edges & & Nodes & Edges\\
    \midrule
    REDDIT-B  & 2 & 1600 & 233 & 274 & 400 & 1212 & 1392 \\
    REDDIT-5K & 5 & 4000 & 375 & 433 & 1000 & 1043 & 1246\\
    REDDIT-12K& 11& 9507 & 258 & 277 & 2390 & 924 & 1066 \\
    Citation  & 2 & 400 & 812 & 898 & 100 & 2886 & 4288\\
    Bitcoin   & 2 & 160 & 412 & 977 & 40 & 880 & 2996 \\
    Q\&A      & 4 & 800 & 918 & 1397 & 200 & 4295 & 5795 \\
    Social    & 2 & 800 & 492 & 458 & 200 & 2713 & 2410 \\
    \bottomrule
  \end{tabular}
\end{table}

All datasets are divided into 80\% small graphs and 20\% large graphs. For the two static network datasets, they are sorted in ascending order of node count, with the top 80\% categorized as small graphs and the bottom 20\% as large graphs, which are then shuffled. The four dynamic graph datasets are sampled based on a set target number of edges, where small and large networks are sampled differently, ensuring no overlap between the two sizes.

For the persistent diagram approximation experiments, the baselines used include PDGNN \cite{neuralapproximation}, TOGL \cite{tgnn}, and RePHINE \cite{beyond}. For the dynamic graph classification experiments, we utilized static graph embedding methods such as GCN \cite{gcn}, GAT \cite{gat}, GraphSage \cite{graphsage}, and GIN \cite{gin}, as well as dynamic graph embedding methods like DHGAT \cite{dhgat}, DySAT \cite{dysat}, Roland \cite{roland}, EGCNO, and EGCNH \cite{evolvegcn}.

\begin{itemize}
  \item \(\mathbf{PDGNN}\) \cite{neuralapproximation}: A method designed to approximate the extended persistent homology of graphs based on the Vietoris-Rips (VP) complexes.
  \item \(\mathbf{TOGL}\) \cite{tgnn}: A novel layer that integrates the persistent homology into Graph Neural Networks (GNNs) based on the Vietoris-Rips (VP) complexes.
  \item \(\mathbf{RePHINE}\) \cite{beyond}: A method that combines vertex- and edge-level PH to create a more expressive topological descriptor, which can be incorporated into GNN layers, enhancing their ability to learn topological features.
  \item \(\mathbf{DySAT}\) \cite{dysat}: A Euclidean dynamic graph embedding approach that employs self-attention mechanisms across both structural and temporal layers.
  \item \(\mathbf{DHGAT}\) \cite{dhgat}: A dynamic hyperbolic graph attention network that utilizes a spatiotemporal self-attention mechanism based on hyperbolic distances.
  \item \(\mathbf{EvolveGCN}\) \cite{evolvegcn}: A Euclidean dynamic graph embedding model that uses GCNs to capture structural information of nodes and RNNs to update the GCN parameters directly. This category includes two architectures: \(\mathbf{EGCN-O}\) and \(\mathbf{EGCN-H}\).
  \item \(\mathbf{Roland}\) \cite{roland}: A Euclidean dynamic graph embedding model that adapts static GNNs for dynamic graphs by treating node embeddings at different GNN layers as hierarchical states and updating them over time. It also approaches the training process as a meta-learning problem for quick adaptation.
\end{itemize}

\subsection{Approximating PD}\label{5.2}
In this subsection, we evaluate the approximation error between the PDs predicted by different methods and the ground truth.

\textbf{Experimental Setup.} Consistent with the setup in \cite{neuralapproximation}, we use the following two metrics to assess the quality of the approximation: (1) The 2-Wasserstein distance between the predicted PD and the ground truth. (2) The total squared distance between the persistence images (PI) converted from the predicted PD and the ground truth, denoted as PIE. Our DNDN method uses the 2-Wasserstein distance(WD) as the loss function, and the suffix \(\_PI\) indicates that the method's loss function is designed based on PIE.

Experiments were conducted on small graph datasets, with the dataset being randomly split into an \(80\%/20\%\) distribution for the training/testing set.  For our Dynamic Neural Dowker Network (DNDN) method, edge features are directly input into the graph as initial features. On static datasets, the edge filter function is designed based on the degree of nodes connected by edges. Concurrently, the line graph transformation generates identical source and sink line graphs. For models that are designed based on graph neural networks, node features result from the aggregation of their edges' features.

\textbf{Results.} As seen in \cref{error:static} and \cref{error:static}, our method exhibits a significant advantage in approximating Dowker persistent homology results, with the best results highlighted in bold. The methods GIN\_PI and GAT\_PI, which use PI as the loss function, do not produce PD outputs, hence the 2-Wasserstein distance cannot be calculated for them. PDGNN, designed to capture EPD, performs suboptimally and lacks support for dynamic directed graph data, leading to poorer performance on some dynamic datasets. Across both static and dynamic datasets, our method significantly outperforms the node-based GNN baselines, demonstrating that our edge-based line graph neural network effectively captures the topological features corresponding to Dowker complexes. 

\begin{table*}[h!]
  \centering
  \caption{Approximation error on static datasets}
  \label{error:static}
  \begin{tabular}{lcccccc}
  \hline
  \multirow{2}{*}{Method}& \multicolumn{2}{c}{REDDIT-B} & \multicolumn{2}{c}{REDDIT-5K} & \multicolumn{2}{c}{REDDIT-12K}\\
  \cmidrule(lr){2-3} \cmidrule(lr){4-5} \cmidrule(lr){6-7}
  & WD & PIE & WD & PIE & WD & PIE \\
  \hline
  GIN\_PI & - & 1.78e-03 $\pm$ 7.0e-04 & - & 2.20e-04 $\pm$ 4.3e-04 & - & 5.18e-04 $\pm$ 4.3e-04 \\
  GAT\_PI & - & 1.57e-03 $\pm$ 3.5e-04 & - & 5.49e-04 $\pm$ 1.7e-04 & - & 7.82e-04 $\pm$ 2.2e-04 \\
  GAT & 0.910 $\pm$ 0.12 & 7.73e-03 $\pm$ 1.0e-02 & 0.731 $\pm$ 0.01 & 5.36e-04 $\pm$ 2.0e-04 & 0.794 $\pm$ 0.01 & 1.48e-03 $\pm$ 3.8e-04 \\
  PDGNN & 0.679 $\pm$ 0.29 & 2.91e-03 $\pm$ 3.0e-03 & 0.697 $\pm$ 0.04 & 5.10e-04 $\pm$ 2.2e-04 & 0.744 $\pm$ 0.03 & 5.10e-04 $\pm$ 2.2e-04 \\
  TOGL & 1.114 $\pm$ 0.19 & 1.82e-03 $\pm$ 5.0e-04 & 0.829 $\pm$ 0.08 & 1.95e-03 $\pm$ 7.3e-04 & 1.021 $\pm$ 0.04 & 1.95e-03 $\pm$ 7.3e-04 \\
  RePHINE & 0.816 $\pm$ 0.01 & 6.78e-04 $\pm$ 1.4e-05 & 0.523 $\pm$ 0.01 & 4.12e-04 $\pm$ 2.5e-04 & 0.685 $\pm$ 0.04 & 4.12e-04 $\pm$ 2.5e-04 \\
  DNDN-EF & 0.610 $\pm$ 0.05 & 7.68e-04 $\pm$ 1.3e-04 & 0.498 $\pm$ 0.07 & 3.78e-04 $\pm$ 4.2e-05 & 0.595 $\pm$ 0.03 & 3.78e-04 $\pm$ 4.2e-05 \\
  DNDN & \textbf{0.499 $\pm$ 0.01} & \textbf{1.56e-04 $\pm$ 1.5e-05} & \textbf{0.317 $\pm$ 0.05} & \textbf{5.21e-05 $\pm$ 1.5e-05} & \textbf{0.389 $\pm$ 0.05} & \textbf{6.73e-05 $\pm$ 1.6e-05} \\
  \hline
  \end{tabular}
  \end{table*}

  \begin{table*}[h!]
    \centering
    \caption{Approximation error on dynamic datasets}
    \label{error:dynamic}
    \setlength{\tabcolsep}{3pt}
    \small
    \begin{tabular}{lcccccccc}
    \hline
    \multirow{2}{*}{Method}& \multicolumn{2}{c}{Citation} & \multicolumn{2}{c}{Q$\&$A} & \multicolumn{2}{c}{Bitcoin} & \multicolumn{2}{c}{Social} \\ 
    \cmidrule(lr){2-3} \cmidrule(lr){4-5} \cmidrule(lr){6-7} \cmidrule(lr){8-9}
    & WD & PIE & WD & PIE & WD & PIE & WD & PIE \\
    \hline
    GIN\_PI & - & 4.71e-04 $\pm$ 1.6e-04 & - & 2.73e-03 $\pm$ 9.1e-04 & - & 5.15e-03 $\pm$ 1.8e-03 & - & 1.44e-03 $\pm$ 1.9e-04 \\
    GAT\_PI & - & 7.82e-04 $\pm$ 2.2e-04 & - & 1.93e-03 $\pm$ 3.6e-04 & - & 2.80e-03 $\pm$ 7.9e-04 & - & 9.04e-04 $\pm$ 1.1e-04 \\
    GAT & 0.960 $\pm$ 0.11 & 1.40e-03 $\pm$ 6.3e-03 & 2.508 $\pm$ 0.11 & 1.09e-01 $\pm$ 1.8e-01 & 3.185 $\pm$ 1.10 & 1.44e-01 $\pm$ 2.4e-01 & 0.900 $\pm$ 0.01 & 9.20e-04 $\pm$ 3.7e-04 \\
    PDGNN & 1.313 $\pm$ 0.44 & 1.87e-02 $\pm$ 2.0e-02 & 2.016 $\pm$ 0.44 & 6.81e-02 $\pm$ 9.4e-02 & 3.708 $\pm$ 1.74 & 4.13e-01 $\pm$ 4.5e-01 & 1.010 $\pm$ 0.16 & 2.34e-03 $\pm$ 3.4e-03 \\
    TOGL & 0.935 $\pm$ 0.07 & 2.45e-03 $\pm$ 2.0e-03 & 1.622 $\pm$ 0.07 & 2.12e-02 $\pm$ 3.3e-02 & 2.064 $\pm$ 0.19 & 8.73e-03 $\pm$ 3.2e-02 & 0.943 $\pm$ 0.04 & 1.54e-03 $\pm$ 1.3e-03 \\
    RePHINE & 0.775 $\pm$ 0.02 & 3.38e-04 $\pm$ 1.5e-04 & 1.867 $\pm$ 0.02 & 2.50e-02 $\pm$ 8.8e-03 & 2.270 $\pm$ 0.01 & 4.88e-02 $\pm$ 2.3e-02 & 0.703 $\pm$ 0.01 & 6.81e-04 $\pm$ 4.6e-04 \\
    DNDN-EF & 0.815 $\pm$ 0.01 & 3.98e-04 $\pm$ 4.7e-05 & 1.364 $\pm$ 0.01 & 8.28e-03 $\pm$ 1.0e-02 & 1.442 $\pm$ 0.23 & 1.22e-02 $\pm$ 1.1e-02 & 0.654 $\pm$ 0.12 & 3.15e-04 $\pm$ 3.1e-04 \\
    DNDN & \textbf{0.591 $\pm$ 0.02} & \textbf{1.29e-04 $\pm$ 3.1e-05} & \textbf{0.804 $\pm$ 0.02} & \textbf{1.33e-03 $\pm$ 7.4e-04} & \textbf{0.908 $\pm$ 0.04} & \textbf{2.13e-03 $\pm$ 2.1e-04} & \textbf{0.514 $\pm$ 0.03} & \textbf{1.01e-04 $\pm$ 1.37e-05} \\
    \hline
    \end{tabular}
    \end{table*}

\subsection{Transferability and Efficiency}\label{5.3}
In this subsection, we designed experiments to investigate the transferability and efficiency of the DNDN algorithm, focusing primarily on its adaptability to real-world large graphs that are computationally expensive to analyze.

\textbf{Experimental Setup.} For transferability, we employed pretrained models obtained from training on small graph datasets within the same category and tested them on large graph datasets. This approach was used to verify DNDN's capability to learn topological features from small graphs of the same category and apply this knowledge to larger graphs. Regarding efficiency, we compared the time taken by DNDN and the GUDHI\cite{GUDHI} method to compute Dowker PDs.

\textbf{Results.} In \cref{Transferability}, "Standard" refers to the results obtained by training directly on large graph datasets, "Pre\_train" denotes the results of testing pretrained models (trained on small graph datasets) on large graph datasets, and "Fine\_tune" represents the outcomes after fine-tuning the pretrained models for 10 epoches. Across all datasets, the performance of models after fine-tuning surpasses that of models trained directly from scratch. This improvement is likely attributed to the fine-tuned models having learned more data information. On dynamic datasets, although "Pre\_train" does not surpass Standard, "fine\_tune" generally yields better results than direct training. These experiments validate our model's transferability to large graphs, which is crucial for addressing the computational expense of persistent homology methods on large dynamic datasets.

In \cref{time}, we compare the efficiency of algorithms in GUDHI and DNDN in processing Dowker PDs on large graph datasets. It is evident that our method significantly outperforms GUDHI on large graph datasets, demonstrating the efficiency and applicability of DNDN in handling complex, large-scale topological data analysis tasks.

\begin{table*}[h!]
  \centering
  \caption{Transferability across graph datasets of varying sizes (WD)}
  \label{Transferability}
  \begin{tabular}{lccccccc}
  \hline
  \multirow{2}{*}{Method} & \multicolumn{3}{c}{Static Network}  & \multicolumn{4}{c}{Dynamic Network}\\
  \cmidrule(lr){2-4} \cmidrule(lr){5-8}
   & REDDIT-B & REDDIT-5K & REDDIT-12K & Citation & Bitcoin & Q \& A & Social \\
  \hline
  Standard & 0.553 $\pm$ 0.01 & 0.488 $\pm$ 0.04 & 0.409 $\pm$ 0.12 & 0.838 $\pm$ 0.05 & 0.849 $\pm$ 0.04 & 1.355 $\pm$ 0.12 & 0.737 $\pm$ 0.01 \\
  Pre\_train & 0.438 $\pm$ 0.01 & 0.177 $\pm$ 0.01 & 0.183 $\pm$ 0.01 & 0.850 $\pm$ 0.05 & 0.924 $\pm$ 0.02 & 1.268 $\pm$ 0.02 & 0.739 $\pm$ 0.01 \\
  Fine\_tune & 0.424 $\pm$ 0.01 & 0.173 $\pm$ 0.01 & 0.176 $\pm$ 0.01 & 0.704 $\pm$ 0.01 & 0.831 $\pm$ 0.01 & 1.121 $\pm$ 0.02 & 0.708 $\pm$ 0.01 \\
  \hline
  \end{tabular}
  \end{table*}

  \begin{table*}[h!]
    \centering
    \caption{Time evaluation on different datasets (seconds)}
    \label{time}
    \begin{tabular}{lccccccc}
    \hline
    \multirow{2}{*}{Method} & \multicolumn{3}{c}{Static Network} & \multicolumn{4}{c}{Dynamic Network} \\
    \cmidrule(lr){2-8}
     & REDDIT-B & REDDIT-5K & REDDIT-12K & Citation & Q \& A & Bitcoin & Social \\
    \hline
    GUDHI & 16.46 $\pm$ 0.001 & 4.64 $\pm$ 0.002 & 5.10 $\pm$ 0.002 & 1.37 $\pm$ 0.001 & 1.33 $\pm$ 0.002 & 3.73 $\pm$ 0.001 & 4.85 $\pm$ 0.002 \\
    DNDN & 1.69 $\pm$ 0.003 & 0.66 $\pm$ 0.003 & 0.70 $\pm$ 0.003 & 0.13 $\pm$ 0.003 & 0.14 $\pm$ 0.003 & 0.17 $\pm$ 0.001 & 0.31 $\pm$ 0.003 \\
    \hline
    \end{tabular}
    \end{table*}

\subsection{Graph Classification}\label{5.4}
In this subsection, having validated DNDN's ability to approximate Dowker PDs and its transferability, we explore the algorithm's accuracy in downstream tasks.

\textbf{Experimental Setup.} To ascertain DNDN's efficacy in graph classification tasks, we orchestrated two experiments: (1) Classification on small graphs, employing 5-fold cross-validation across each dataset for final outcome derivation. (2) Transferable classification experiment, where the model, initially trained on small graphs, was subjected to graph classification tasks on larger graphs. All experimental approaches leveraged the mean-pooling method to derive graph embeddings for label prediction.

\textbf{Results.} As indicated in \cref{accuracy}, 'Small' denotes the classification experiments conducted directly on small graph datasets, whereas 'Large' refers to graph classification on larger datasets after training on small graph datasets. Dynamic network embedding methods like DySAT, EGCNO, and DHGAT, which focus on capturing individual node's evolving patterns, tend to lose some evolutionary information after mean-pooling. Conversely, our method achieved optimal performance in most scenarios, primarily because it learns Dowker PDs that represent the graph's topological features, thus capturing the graph's global characteristics. Most baseline methods are designed around a node's neighborhood, failing to adequately capture the graph's comprehensive information. Particularly in the transferable classification experiments, DNDN outperformed across mostly datasets, demonstrating that insights learned from small graph datasets can be effectively transferred to larger graph datasets.

\begin{table*}[h!]
  \centering
  \caption{Accuracy on graph classification task}
  \label{accuracy}
  \setlength{\tabcolsep}{3pt}
  \small
  \begin{tabular}{lcccccccccccc}
  \hline
  \multirow{3}{*}{Method} & \multicolumn{4}{c}{Static Network} & \multicolumn{8}{c}{Dynamic Network} \\
  \cmidrule(lr){2-13} 
   & \multicolumn{2}{c}{REDDIT-B} & \multicolumn{2}{c}{REDDIT-5K} &  \multicolumn{2}{c}{Citation} & \multicolumn{2}{c}{Q\&A} & \multicolumn{2}{c}{Bitcoin} & \multicolumn{2}{c}{Social} \\
   \cmidrule(lr){2-3} \cmidrule(lr){4-5} \cmidrule(lr){6-7} \cmidrule(lr){8-9}\cmidrule(lr){10-11} \cmidrule(lr){12-13}   
   & Small & Large & Small & Large & Small & Large & Small & Large & Small & Large & Small & Large  \\
  \hline
  GCN & 73.8 $\pm$ 0.5 & 50.1 $\pm$ 0.1 & 33.0 $\pm$ 0.3 & 20.7 $\pm$ 0.0 & 50.0 $\pm$ 0.0 & 52.5 $\pm$ 0.7 & 65.0 $\pm$ 0.8 & 49.5 $\pm$ 0.0 & 84.4 $\pm$ 1.1 & 52.3 $\pm$ 0.3 & 84.0 $\pm$ 0.6 & 88.2 $\pm$ 0.2 \\
  GAT & 79.4 $\pm$ 1.2 & 52.1 $\pm$ 0.0 & 38.1 $\pm$ 0.2 & 20.0 $\pm$ 0.0 & 51.2 $\pm$ 0.1 & 49.0 $\pm$ 0.1 & 41.9 $\pm$ 0.2 & 57.0 $\pm$ 0.4 & 87.5 $\pm$ 1.2 & 50.1 $\pm$ 0.3 & 71.3 $\pm$ 1.2 & 81.3 $\pm$ 0.6 \\
  GraphSage & 79.1 $\pm$ 0.4 & 52.3 $\pm$ 0.2 & 33.2 $\pm$ 0.2 & 21.5 $\pm$ 0.0 & 67.5 $\pm$ 0.3 & 56.0 $\pm$ 0.1 & 76.3 $\pm$ 0.7 & 53.0 $\pm$ 0.4 & 78.1 $\pm$ 0.1 & 62.3 $\pm$ 0.3 & 91.8 $\pm$ 1.5 & 89.5 $\pm$ 1.2 \\
  GIN & 73.8 $\pm$ 0.2 & 51.1 $\pm$ 0.0 & 20.9 $\pm$ 0.0 & 25.5 $\pm$ 0.1 & 83.8 $\pm$ 0.3 & 51.1 $\pm$ 0.6 & 70.0 $\pm$ 1.2 & 68.0 $\pm$ 1.8 & 68.8 $\pm$ 1.3 & 71.2 $\pm$ 1.2 & 81.3 $\pm$ 0.6 & 78.5 $\pm$ 0.6 \\
  DySAT & 78.3 $\pm$ 0.7 & 54.5 $\pm$ 0.0 & 32.2 $\pm$ 0.1 & 22.3 $\pm$ 0.0 & 86.9 $\pm$ 1.3 & 63.0 $\pm$ 1.6 & 78.4 $\pm$ 2.5 & 77.3 $\pm$ 2.7 & 87.4 $\pm$ 1.4 & 72.2 $\pm$ 0.7 & 91.2 $\pm$ 1.5 & 75.2 $\pm$ 2.0 \\
  DHGAT & 76.4 $\pm$ 0.1 & 56.2 $\pm$ 0.1 & 36.7 $\pm$ 0.2 & 21.7 $\pm$ 0.0 & 85.8 $\pm$ 1.2 & 65.9 $\pm$ 0.3 & 76.7 $\pm$ 0.1 & 78.4 $\pm$ 0.2 & 86.4 $\pm$ 0.3 & 65.9 $\pm$ 0.3 & 92.3 $\pm$ 0.4 & 86.5 $\pm$ 0.2 \\
  EGCN-O & 82.2 $\pm$ 0.4 & 61.0 $\pm$ 0.0 & 34.8 $\pm$ 0.1 & 30.1 $\pm$ 0.1 & 84.5 $\pm$ 0.8 & 63.4 $\pm$ 0.3 & 79.2 $\pm$ 0.4 & 78.3 $\pm$ 0.6 & 89.3 $\pm$ 0.6 & 73.3 $\pm$ 0.2 & 89.2 $\pm$ 0.4 & 81.2 $\pm$ 0.9 \\
  EGCN-H & 82.5 $\pm$ 0.5 & 59.5 $\pm$ 0.2 & 32.5 $\pm$ 0.0 & 28.3 $\pm$ 0.0 & 86.7 $\pm$ 0.2 & 62.4 $\pm$ 0.5 & 80.4 $\pm$ 0.5 & 80.2 $\pm$ 0.6 & \textbf{89.6 $\pm$ 0.4} & 75.5 $\pm$ 0.6 & 88.6 $\pm$ 1.2 & 80.5 $\pm$ 1.4 \\
  Roland & \textbf{84.6 $\pm$ 0.2} & 65.2 $\pm$ 0.2 & 47.2 $\pm$ 0.1 & 40.2 $\pm$ 0.2 & \textbf{87.5 $\pm$ 0.2} & 68.2 $\pm$ 0.7 & 78.5 $\pm$ 0.6 & 78.6 $\pm$ 0.4 & 80.2 $\pm$ 0.2 & 72.1 $\pm$ 0.5 & 90.5 $\pm$ 0.2 & \textbf{91.2 $\pm$ 0.3} \\
  DNDN & 83.4 $\pm$ 0.2 & \textbf{73.6 $\pm$ 1.2} & \textbf{56.7 $\pm$ 0.4} & \textbf{41.3 $\pm$ 1.0} & 85.6 $\pm$ 0.2 & \textbf{72.4 $\pm$ 0.4} & \textbf{83.4 $\pm$ 0.4} & \textbf{81.2 $\pm$ 0.5} & 84.5 $\pm$ 0.5 & \textbf{77.6 $\pm$ 1.0} & \textbf{94.5 $\pm$ 0.2} & 89.2 $\pm$ 0.1 \\
  \hline
  \end{tabular}
  \end{table*}

\section{conclusion}
In conclusion, our study introduces the Dynamic Neural Dowker Network (DNDN), a pioneering framework designed to tackle the computational complexities associated with applying persistent homology to dynamic graphs. Through a novel integration of line graph transformations and a Source-Sink Line Graph Neural Network (SSLGNN), coupled with a duality edge fusion mechanism, DNDN effectively captures and approximates the Dowker persistent homology results of dynamic networks. Our experimental evaluation, spanning both static and dynamic datasets, demonstrates DNDN's superior performance in approximating true persistent homology, highlighting its potential to enhance graph classification tasks. Notably, the method showcases remarkable transferability, proving its efficacy in learning topological features from smaller graphs and applying them to larger counterparts with enhanced efficiency. In the future, we will focus on two critical directions to address the current limitations of DNDN: (1) extending the dynamic Dowker filtration method to node-level tasks by constructing dynamic neighborhood subgraphs of nodes to study their higher-order evolutionary patterns, and (2) building on the approximation of 0-dimensional and 1-dimensional persistence diagrams (PDs), exploring methods to approximate higher-dimensional PDs.

\begin{acks}
  This work was supported in part by the National Natural Science Foundation of China under Grant U19B2004.
\end{acks}

\bibliographystyle{ACM-Reference-Format}
\bibliography{sample-base}

\appendix

\section{Dowker Complexes}
\subsection{Diffences between VP complexes and Dowker complexes}

\begin{figure}[ht]
  \centering
  \includegraphics[width=\columnwidth]{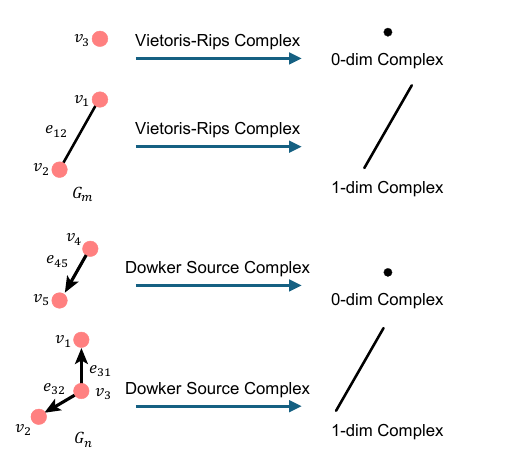}
  \caption{An example demonstrating the difference Vietoris-Rips (VP) complexes and Dowker complexes.}
  \label{figaa}
\end{figure}

As \cref{figaa} illustrated, Vietoris-Rips (VP) complexes and Dowker complexes focus on different graph structures. VP complexes directly concern the neighbor structures on the graph, whereas Dowker complexes are interested in the shared neighbor structures on the graph. In graph \(G_m\), \(v_1\) and \(v_2\) are connected by edge \(e_{12}\), corresponding to a one-dimensional complex. In graph \(G_n\), \(v_1\) and \(v_2\) correspond to the same one-dimensional complex because \(e_{31}\) and \(e_{32}\) share a common source node \(v_3\), even though \(v_1\) and \(v_2\) may not be directly connected by an edge. Traditional graph neural networks aggregate the neighbors of a node, making them suitable for VP complexes but not necessarily for Dowker complexes. However, we find that the relationship between \(e_{31}\) and \(e_{32}\) can be transformed into the form of a line graph, thus adapting line graph neural networks for computation with Dowker complexes.

\section{pseudocode}
In this section, we introduce the pseudocode for constructing source line graphs and sink line graphs, referenced as \cref{alg1}, and the pseudocode describing the DNDN model, cited as \cref{alg2}.

\begin{algorithm}
  \SetAlgoLined
  \KwIn{Directed graph \(\mathcal{G}_d=(\mathcal{V},\mathcal{E} )\), where \(\mathcal{V}\) is the set of vertices and \(\mathcal{E}\) is the set of directed edges}
  \KwOut{Source line graph \(L(\mathcal{G}_d)^{so}=(\mathcal{V}_{L(\mathcal{G}_d)}^{so}, \mathcal{E}_{L(\mathcal{G}_d)}^{so})\) and Sink line graph \(L(\mathcal{G}_d)^{si}=(\mathcal{V}_{L(\mathcal{G}_d)}^{si}, \mathcal{E}_{L(\mathcal{G}_d)}^{si})\)}
  \SetKwFunction{FMain}{ConstructLineGraphs}
  \SetKwProg{Fn}{Function}{:}{}
  \Fn{\FMain{\(G\)}}{
      Initialize \(\mathcal{V}_{L(\mathcal{G}_d)}^{so} \leftarrow \emptyset\)\;
      Initialize \(\mathcal{E}_{L(\mathcal{G}_d)}^{so} \leftarrow \emptyset\)\;
      Initialize \(\mathcal{V}_{L(\mathcal{G}_d)}^{si} \leftarrow \emptyset\)\;
      Initialize \(\mathcal{E}_{L(\mathcal{G}_d)}^{si} \leftarrow \emptyset\)\;
      
      \tcc{Constructing the Source Line Graph}
      \ForEach{edge \(e = (u, v) \in \mathcal{E}\)}{
          Add \(e\) to \(\mathcal{V}_{L(\mathcal{G}_d)}^{so}\)\; 
      }
      \ForEach{pair of edges \(e_1 = (u, v)\) and \(e_2 = (u, w) \in \mathcal{E}\)}{
          \If{\(e_1 \neq e_2\)}{
              Add edge \((e_1, e_2)\) to \(\mathcal{E}_{L(\mathcal{G}_d)}^{so}\)\; 
          }
      }
      
      \tcc{Constructing the Sink Line Graph}
      \ForEach{edge \(e = (u, v) \in \mathcal{E}\)}{
          Add \(e\) to \(\mathcal{V}_{L(\mathcal{G}_d)}^{si}\)\; 
      }
      \ForEach{pair of edges \(e_1 = (u, v)\) and \(e_2 = (w, v) \in \mathcal{E}\)}{
          \If{\(e_1 \neq e_2\)}{
              Add edge \((e_1, e_2)\) to \(\mathcal{E}_{L(\mathcal{G}_d)}^{si}\)\;
          }
      }
      
      \Return{\(L(\mathcal{G}_d)^{so}=(\mathcal{V}_{L(\mathcal{G}_d)}^{so}, \mathcal{E}_{L(\mathcal{G}_d)}^{so})\) , \(L(\mathcal{G}_d)^{si}=(\mathcal{V}_{L(\mathcal{G}_d)}^{si}, \mathcal{E}_{L(\mathcal{G}_d)}^{si})\)}\;
  }
  \caption{Algorithm to construct the source and sink line graphs from a directed graph \(\mathcal{G}_d\)}
  \label{alg1}
  \end{algorithm}

\afterpage{
\begin{algorithm}
  \SetAlgoLined
  \SetNlSty{textbf}{[}{]}
  \DontPrintSemicolon
  \KwIn{Source line graph \(L(\mathcal{G}_d)^{so}=(\mathcal{V}_{L(\mathcal{G}_d)}^{so}, \mathcal{E}_{L(\mathcal{G}_d)}^{so})\); sink line graph \(L(\mathcal{G}_d)^{si}=(\mathcal{V}_{L(\mathcal{G}_d)}^{si}, \mathcal{E}_{L(\mathcal{G}_d)}^{si})\); dynamic Dowker filtration \(\mathcal{W}\);aggregation function \(AGG\); message-passing function \(MSG\); Edge Fusion function \(EdgeFusion\); neighborhood function \(\mathcal{N}\); multilayer perceptron function \(MLP\); total number of layers in the network \(M\)}

  \ForEach{ \(u \in \mathcal{V}_{L(\mathcal{G}_d)}^{so}=\mathcal{V}_{L(\mathcal{G}_d)}^{si}\)}{
    \(h^0_{u_{so}} \leftarrow \mathcal{W}(u)\)\;
    \(h^0_{u_{si}} \leftarrow \mathcal{W}(u)\)\;
  }
  
  \For{each layer \(m = 1\) \KwTo \(M\)}{
      \ForEach{ \(u \in \mathcal{V}_{L(\mathcal{G}_d)}^{so}=\mathcal{V}_{L(\mathcal{G}_d)}^{si}\)}{
          Calculate \( h_{u_{so}}^m \) for source line graph:
          \(h_{u_{so}}^m \leftarrow AGG^m\left(\left\{MSG^m\left(h_{v_{ef}}^{m-1}\right), v_{so} \in \mathcal{N}(u_{so})\right\}, h_{u_{ef}}^{m-1}\right)\)\;
          Calculate \(h_{u_{si}}^m\) for sink line graph:
          \(h_{u_{si}}^m \leftarrow AGG^m\left(\left\{MSG^m\left(h_{v_{ef}}^{m-1}\right), v_{si} \in \mathcal{N}(u_{si})\right\}, h_{u_{ef}}^{m-1}\right)\)\;
          Perform edge fusion to update features:
          \(h_{u_{ef}}^m \leftarrow EdgeFusion \left( h_{u_{so}}^m, h_{u_{si}}^m\right)\)\;
      }
  }
  \(\mathbf{H^n} \leftarrow \left\{ h^n_{u_{ef}} : u \in \mathcal{V}_{L(\mathcal{G}_d)}^{so}=\mathcal{V}_{L(\mathcal{G}_d)}^{si} \right\}\)\;
  \(PD^0 \leftarrow MLP(\mathbf{H^n})\)\;

  \(h^n_{\text{agg}} \leftarrow \sum_{v \in \mathcal{N}(u_{si}) \bigcup \mathcal{N}(u_{so})} \mathcal{W}(v) \cdot h^n_v\)\;
  \(\mathbf{H}^n_{\text{agg}} \leftarrow \left\{ h^n_{\text{agg}}\right\}\)\;
  \(PD^1 \leftarrow MLP(\mathbf{H}^n_{\text{agg}})\)\;
  \caption{Dynamic Neural Dowker Network (DNDN) Forward Propagation Algorithm}
  \label{alg2}
  \end{algorithm}
}

\end{document}